\documentclass{article}

 \usepackage[dblblindworkshop, final]{neurips_2026}
\workshoptitle{AI and the Self}

\usepackage[utf8]{inputenc} 
\usepackage[T1]{fontenc}    
\usepackage{hyperref}       
\usepackage{url}            
\usepackage{booktabs}       
\usepackage{amsfonts}       
\usepackage{nicefrac}       
\usepackage{microtype}      
\usepackage{xcolor}         
\usepackage{multirow}
\usepackage{graphicx}
\usepackage{float}
\usepackage{array}
\usepackage{tcolorbox}
\usepackage{fontawesome5}

\newtcolorbox{takeaway}{
    colback=yellow!10,
    colframe=yellow!50!black,
    boxrule=0.8pt,
    arc=1pt,
    left=2pt, right=2pt, top=2pt, bottom=2pt,
}
\hypersetup{hidelinks}

\title{Ownership in AI-Assisted Everyday Tasks}

\author{
  Megan Wei$^{1}$ \And
  Melanie Subbiah$^{1}$ \And
  Audrey Lee$^{2}$ \And
  Annya Dahmani$^{3}$ \AND
  Dave Edwards$^{4}$ \And
  Helen Edwards$^{4}$ \And
  Ellie Pavlick$^{1}$ \AND
  \\
  $^{1}$Brown University \quad \texttt{\{meganwei, m.subbiah, ellie\_pavlick\}@brown.edu} \\
  $^{2}$Carnegie Mellon University \quad \texttt{ael2@andrew.cmu.edu} \\
  $^{3}$University of California, Berkeley \quad \texttt{adahmani@berkeley.edu} \\
  $^{4}$Artificiality Institute \quad \texttt{\{dave, helen\}@artificialityinstitute.org}
}

\begin{document}

\maketitle

\begin{abstract}
  When does work done with AI still feel like ours? As AI becomes woven into everyday tasks, we must examine what happens to our sense of ownership and contribution when a machine shares in producing what we make. We report an exploratory qualitative survey in which participants were asked to describe two recent, self-selected tasks completed with AI: one that felt like their own and one that did not. We find that felt ownership depends on the process of collaboration: people disown work when they merely approve AI's suggestions, but retain ownership when they lead, iterate, or rewrite. Ownership can also extend to settings where people own the vision for a project but not the execution; respondents reported high ownership on tasks they could not have completed without AI.  Loss of personal voice and a lack of comprehension of the output both erode ownership. Finally, willingness to disclose AI use is often decoupled from actual pride or ownership, and instead shaped by community norms and fear of credit erasure. We propose several research directions as a result of these findings to promote AI development that supports people's sense of authorship over their own lives.
\end{abstract}

\section{Introduction}

Artificial Intelligence (AI) is rapidly becoming integrated into both our personal and work lives, primarily through Large Language Models (LLMs) used as chatbots or agents \citep{HOSSEINI2025100399}. It is, therefore, imperative to consider the effect this integration will have not just on our work, but on our personal well-being \citep{sharma2026s, yang2026ai, luettgau2025people}. The relevant research and media attention have largely focused on the most high-stakes areas of our well-being, involving self-harm, death, and crises of mental health \citep{guo2024large, bernier2026mass, archiwaranguprok2025simulating}. However, other important aspects of our well-being that may erode more slowly over time are no less consequential. One such area is our sense of purpose and fulfillment in our life and work, which hinges on some degree of pride or ownership in the decisions and actions we take \citep{doi:10.1037/a0017152, kirk2015m, creativeownership}. In this paper, informed by a qualitative survey, we narrow in on this issue, asking, \textit{How does using AI in everyday tasks affect our sense of ownership of our daily life decisions and work?}

Prior work has defined \textit{cognitive offloading} \citep{risko2016cognitive, soc15010006, grinschgl2021consequences, guingrich2026belief, yang2026ai} as delegating cognitive effort to another entity, such as an AI tool, and thus reducing the user's effort and involvement in the process and output. Cognitive offloading affects comprehension and learning for the person doing the task, often with minimal gains in efficiency \citep{kosmyna2025brainchatgptaccumulationcognitive, yu2026cognitive}. In our work, we focus on the feelings attached to this phenomenon, developing hypotheses for factors that contribute to people feeling ownership of everyday tasks they complete with AI assistance.

We conduct an informal survey\footnote{Participants were recruited through our professional networks, spanning students at research universities and adults in the design and entrepreneurship community, who were generally heavy users of AI tools. We acknowledge this is a biased sample; we report these results as a necessary first step toward surfacing problems and insights in this space, which we encourage future work to validate through more structured research studies. View survey results here: \href{https://aiownership.github.io}{aiownership.github.io}} comparing people's patterns and feelings when using AI in a task over which they ultimately feel high ownership versus a task over which they feel low ownership. Comparing points along this spectrum is in line with \citet{creativeownership}'s work on felt ownership in creative work and allows us to probe differences in the tasks themselves, in interaction patterns with the AI, and in personal reactions to the task and process. Our qualitative analysis of these responses reveals several themes driving ownership: 1) the importance of \textbf{process over output}, 2) a feeling of \textbf{differentiating one's personal voice}, 3) \textbf{achieving something beyond one's own abilities}, and 4) \textbf{comprehension of the output}. Additionally, we uncover that people's \textbf{willingness to disclose AI use} to colleagues or peers is less connected to their actual sense of ownership or pride in their work and \textbf{more contingent on the norms and stigmas in their communities}.


\begin{figure}[H]
    \centering
    \includegraphics[width=0.95\textwidth]{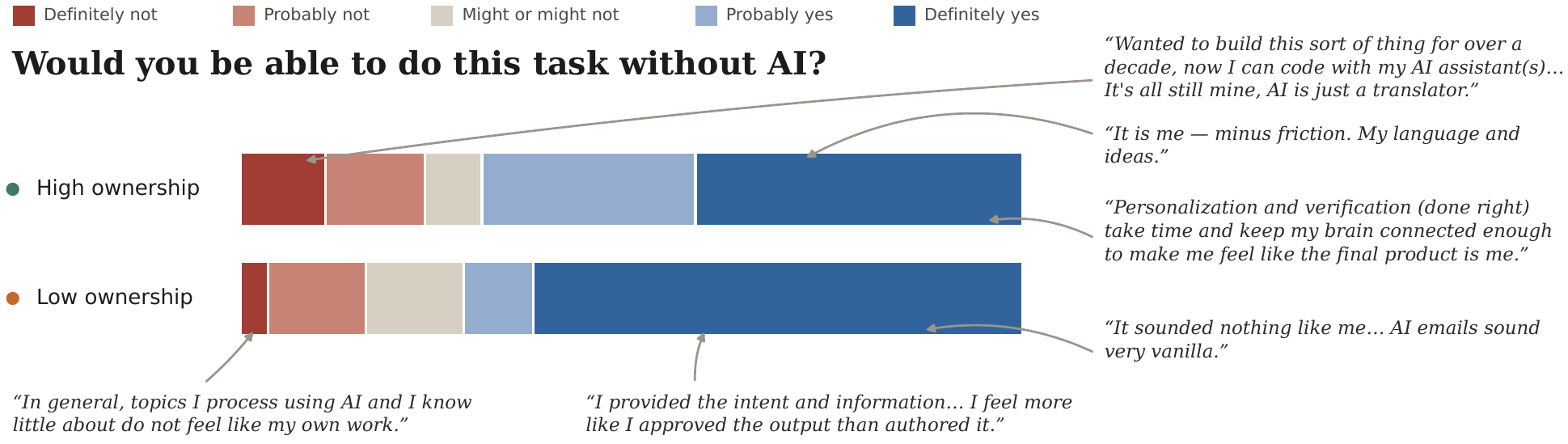}
    \caption{Responses when asked whether users are able to do the task without AI, across high and low ownership tasks.}
    \label{fig:front}
\end{figure}
\section{Survey Design and Methods}

We design a survey (all questions shown in Appendix~\ref{app:survey}) with 1) demographics, 2) general AI usage, 3) two task reflections, 4) a checklist of common AI tasks, and 5) comparative questions. For the task reflections \citep{creativeownership}, participants described one high-ownership and one low-ownership task, shown in randomized order. They answered whether the task would have been done, and \textit{could} have been done, without AI, alongside free-text reflections on identity, independent contribution, pride, and disclosure. Participants selected tasks done using AI in the past 30 days from a list adapted from \citet{chatterji2025chatgpt}. The comparative questions (\textit{Why these tasks? What made the biggest difference in felt ownership?}) assessed differences across tasks \citep{aimemorygap}.

We recruited respondents via professional networks and social media (LinkedIn, Instagram), as well as university mailing lists and Slack channels, collecting responses from August 6-27, 2026. The sample is biased (Appendix \ref{app:demographics}), especially toward mid-to-late career highly educated white adults who heavily use AI and are able to pay out of pocket for it. In total, 117 people began the survey, with 52 complete responses and 64 responses containing at least one task reflection. 

\section{Results}
\paragraph{Process drives ownership.}
When asked what made the difference in ownership between their two tasks, respondents highlighted that ownership was 
about who led the process (\textit{``ownership comes from owning the decisions made during the processing of the information. If there was no processing of information nor synthesis then it doesn't feel I own it.''}). This finding is in line with \citet{creativeownership}'s work on creative ownership which showed \textit{process}, and particularly \textit{control over decisions}, to be a differentiating factor in ownership. 
We also see ownership diminishes when the human simply approves the AI output (\textit{``I reviewed, I looked, I was in the loop, but I wasn't in charge.''}) \citep{dhillon}. 
Some users are intentional about protecting their process to preserve ownership (\textit{``
I augment my own expertise and skills with AI, I don't hand off the whole thing.''}). In collaborative settings, ownership on the shared product shifts based on observations of collaborators' AI usage that may be misaligned with an individual's process (\textit{``Early on, I was more proud of the project… As the deadline approached, I saw my students use AI more and more extensively for filling in gaps.''}).
\vspace{-0.2cm}
\paragraph{Homogeneous voice, personalization, and rejection.}
For writing tasks, a prominent source of disownership was the voice \citep{ippolito2022creativewritingaipoweredwriting}: \textit{``it sounded nothing like me… AI emails sound very vanilla.''}; \textit{``I didn't like how `artificial' it sounded.''}; and \textit{``the sentence structure and big words used did not sound like any human personality''} \citep{idiosyncrasies}, which aligns with findings of homogenization and reduced diversity in AI-generated content \citep{doshihauser, si2025can, homogenousanderson, sourati2026shrinking}. Respondents who invested in personalization \citep{diverseaipersona, 80me20ai, shaikh2025aligning} of models reported retaining ownership (\textit{``consistent with my voice after much training of this project''}; \textit{``My model already has md file for content creation. This shows samples of my work, direct instructions about my tone, guardrails of what not to say or do''}). However, there is a ceiling to this personalization \citep{wang-etal-2025-catch}, where the outputs were \textit{``over aggregated and without some of my unique POV''}.

Interestingly, AI's bad outputs sometimes helped people build ownership by giving them a point of contrast: \textit{``It helped me understand what I didn't want so I could create what I wanted… I definitely felt more ownership… because AI failed so horribly.''}; \textit{``I don't always know what I want… until I see where the AI-generated version misses… even an imperfect output can be useful because the process of disagreeing with it… becomes part of my own thinking.''} This phenomenon has also been evidenced in \citet{aliciacreative}, where writers perceived bad outputs as useful ``anti-patterns'': seeing what they don't want helps them understand what they do want.

\begin{takeaway}
\textcolor{yellow!50!black}{\faLightbulb}\ \textit{Encountering and rejecting bad AI outputs can help people define and own their work.}
\end{takeaway}

\vspace{-0.2cm}
\paragraph{Enabling outputs beyond one's own abilities.}
Counterintuitively, respondents were less certain they could have done their high-ownership task without AI, where 42\% answered ``definitely yes'' against 63\% for low-ownership tasks (Figure~\ref{fig:front}). Some of the most-owned projects would have been impossible for their creators and unlocked new creative powers. For example, two respondents had longed to produce children's books and graphic novels but could not illustrate their writing, with one commenting, \textit{``my inability to get [my vision] out of my head originally led me to stop writing children's books.''} AI helped them realize their envisioned illustrations with great positive effect: \textit{``I feel like I'm truly honoring my younger self''} \citep{novicemusic, child, cococo}. When the human directs the project's vision, AI occupies the role of a hired illustrator or junior teammate (\textit{``it felt like directing a human paralegal. That's my job.''}) \citep{metaphor}, which does not threaten ownership \citep{maier}.

\begin{takeaway}
\textcolor{yellow!50!black}{\faLightbulb}\ \textit{People can feel high ownership over tasks they do not possess the skills to complete without AI.}
\end{takeaway}

Meanwhile, disowned tasks were more often things people could do, but chose to delegate: routine emails, lookups, and formatting. Owned and disowned tasks differ in nature: \textit{``a task that I wanted to have ownership of''} compared to \textit{``a task that I would have rather not have had to do in the first place''}, which aligns with \citet{pierce2001}'s defined motives towards psychological ownership.

\vspace{-0.2cm}

\paragraph{No ownership without comprehension.} 
One route towards psychological ownership \citep{pierce2001} is intimately knowing the task. In our study, ownership failed when respondents could not understand the artifact (\textit{``I really don't understand much of the codebase… I wouldn't consider this my work at all… honestly I'll probably be annoyed if I have to patch it in the future since I don't even feel responsible for it.''}) \citep{copilot, widegap} or evaluate the output (\textit{``I certainly don't know enough about the topic to judge and evaluate the output — hence does not feel like something I would call `mine'''}). Without comprehension, people felt little responsibility for maintaining and defending their work \citep{seocode}. 
\vspace{-0.3cm}
\paragraph{Disclosure.}
Respondents' willingness to disclose AI use varied widely and often depended on the situation. Individuals most willing to disclose AI use had concerns about correctly attributing the source of the work (\textit{``I'd absolutely mention that almost all of it was AI generated, otherwise I'd be taking way more credit than I deserve''}) \citep{fangdisclosure, walters2023fabrication, topaz2026fabricated}. People cautiously open to disclosure generally preferred to narrate the process of AI usage rather than a label of AI-generated or not \citep{hedisclosure}: \textit{``I'd be willing to say I used AI for it if I clarified the limited extent to which it was used.''} Often, they fear credit erasure \citep{reif2025social} under a simple label, worrying people would think \textit{``that I didn't do any work… as if I don't think for myself''} or that \textit{``the code itself was mostly or entirely AI generated, thus discounting my contribution.''} This concern has been reflected in previous work, with AI disclosure eroding trust \citep{SCHILKE2025104405}; damaging perceptions of the creator \citep{raedisclosure}; and impacting marginalized groups, future evaluations, and hiring outcomes \citep{kadomadisclosure}. These concerns drove reluctance for respondents least willing to disclose AI use (\textit{``I don't tend to tell people I use AI for my emails when I do use it''}; \textit{``I would probably be ashamed to admit using the tools''}). Respondents also reported that disclosure depends \textit{``somewhat on the audience and the stakes''}, as prior work found with AI-written dating profiles \citep{datingdisclosure}. The surrounding organizations often influence these audiences: from a company that \textit{``encourages its use,''} to situations where people would disclose to colleagues but not clients, to communities that might assume \textit{``delegation rather than oversight.''} One lecturer reported that they would not share a failed AI experiment with colleagues who \textit{``are in love with AI and won't hear anything bad against it.''} As a result of these complex social pressures, willingness to disclose AI use is not necessarily coupled with feelings of pride or ownership of the work: \textit{``AI helped me clarify my thinking… [but] I think I would have to be careful disclosing that I used AI. Not that I'm not proud, but there is a stigma.''}

\begin{takeaway}
\textcolor{yellow!50!black}{\faLightbulb}\ \textit{Disclosure is a distinct dimension: willingness to disclose AI use does not necessarily accompany pride or ownership of the work.}
\end{takeaway}

\section{Takeaways and Recommendations}
From our survey, ownership emerges as a function of the process: iterating with the model, authoring the prompts, personalizing models, and rejecting and learning from bad outputs. AI has made previously unachievable tasks achievable, and it has created situations where people produce output they can't fully evaluate or explain. As AI becomes more deeply integrated into society \citep{kobiella}, we need to investigate how to preserve people's creative agency \citep{kumarcreative, aliciacreative}, ownership \citep{creativeownership}, and learning across various collaborative, evaluative, and social settings. These findings motivate further research into the themes identified in our results:
\vspace{-0.3cm}
\paragraph{Focus on process.} We encourage study of not just how we can increase human investment in producing AI-assisted work, but also how we can increase process visibility for users and collaborators. For example, does recording and displaying the interaction history (edits, rejections, prompts, decisions, leadership) strengthen felt ownership and calibrate self-attribution? How does visibility of a collaborator's process reshape ownership in shared projects? Can we measure adverse effects when people do not take ownership of the process, such as increased effort for reviewers of the work?
\vspace{-0.3cm}
\paragraph{Personalization.} We encourage future work on personalization to not be purely driven by commercial incentives but also to promote user well-being and identity, such as by investigating the effects of perceived versus actual personalization. Several respondents felt they had uniquely scaffolded their AI to adapt to them, making it now \textit{their AI}. Are these personalization methods actually unique and effective? How does informing people that their scaffolding methods are actually similar change their perception of ownership? Can we measure how differentiated a model is in its responses, and does this influence the user's ownership or identification with the model?
\vspace{-0.3cm}
\paragraph{Enabling the previously impossible.} Our exploration shows that people can feel great ownership over outputs that they could not have produced on their own, like illustrating a graphic novel. However, it is not clear how this sense of ownership coexists with low expertise in some areas of the output. Is this effect primarily present in cases where users may be strong \textit{evaluators} but not strong \textit{producers} of the work (e.g., the art critic vs. the artist)? Does revealing a skilled critique from an expert human collaborator change the felt ownership of an accomplishment outside of one's domain?
\vspace{-0.3cm}
\paragraph{Promoting honest disclosure of AI use.} People can more authentically own their work and learn healthy modes of human-AI collaboration when community cultures encourage honesty around discussing AI use. We encourage not just studying how readers or reviewers of work perceive AI disclosures, but also what factors encourage or discourage the producer from honestly disclosing AI use when it is required. For example, what formats of disclosure are most effective (e.g., categorical vs. narrative)? How does social feedback on output quality affect willingness to disclose use?

\begin{ack}
This material is based upon work supported by the U.S. National Science Foundation (NSF) under Cooperative Agreement No. 2433429, "NSF Al Research Institute on Interaction for Al Assistants (ARIA)”, and by Google LLC.
\end{ack}

\bibliographystyle{plainnat}
\bibliography{main}


\appendix

\section{Survey}
\label{app:survey}
 
We include our Qualtrics survey questions. Response formats are marked MC (single choice), SA (select all that apply), DD (dropdown), and FF (free text). Conditional display logic and the task-reflection section are noted where they apply. Before beginning the survey, participants reviewed a consent page: ``I am 18 or older and I consent to take part in this survey. [MC: Yes, I consent / No].''
  
\subsection{Demographic information}
 
\begin{itemize}
\item What is your age? [DD: Under 18; 18--24; 25--34; 35--44; 45--54; 55--64; 65 or older]
\item In which country do you currently live? [DD: country list]
\item How do you describe your gender? [MC: Woman; Man; Non-binary; Prefer to self-describe (FF); Prefer not to say]
\item How do you describe your race or ethnicity? [SA: Native American; Asian; African American; Hispanic or Latino; Middle Eastern; Pacific Islander; Caucasian; Other (FF); Prefer not to say]
\item What is the highest level of education you have completed? [DD: Less than high school; High school / GED; Some college; Associate's degree; Bachelor's degree; Master's degree; Doctorate / Professional degree; Prefer not to say]
\item What best describes your current employment status? [DD: Employed full-time; Employed part-time; Self-employed; Student; Not currently employed; Retired; Prefer not to say]
\item Which best describes the industry you work in? (If not currently working, select ``Not applicable.'') [DD: Healthcare; Education; Technology / software; Finance / insurance; Retail / hospitality; Manufacturing / trades; Government / public sector; Media / arts / design; Legal; Science / engineering; Transportation / logistics; Other; Not applicable]
\item What is your job title or main role? [FF]
\end{itemize}
 
\subsection{AI usage}
 
\begin{itemize}
\item In the past 3 months, how often have you used any AI chatbot or assistant (e.g., ChatGPT, Gemini, Claude, Copilot, or an AI voice assistant)? [DD: Never; Less than monthly; Monthly; Weekly; Several times a week; Daily; Many times a day]
\item Which AI tools do you use? [SA: ChatGPT (OpenAI); Google Gemini; Claude (Anthropic); Grok (xAI); Microsoft Copilot; Meta AI; Perplexity; DeepSeek; An AI image generator/editor (e.g., Midjourney, Nano Banana); An AI video generator/editor (e.g., Google Veo, Runway); An AI music or audio generator/editor (e.g., Suno); An AI voice assistant or voice mode (e.g., ChatGPT voice, Gemini Live, Siri, Alexa); An AI coding assistant (e.g., Cursor, Copilot); An AI companion/character app (e.g., Character.AI, Replika); An AI chat embedded in a different tool/website (please specify what tool) (FF); Another AI tool (FF)]
\item When you use AI, how do you interact with the AI? [SA: Typing; Voice; Image/video upload; Document/file upload; Tool call/app integrations/MCPs; Other (FF)]
\item When did you start using AI tools regularly? [DD: I don't use them regularly; In the last 3 months; 3--6 months ago; 6--12 months ago; 1--2 years ago; More than 2 years ago]
\item If you are employed, how much do you personally use AI in your job? [DD: Not at all; Rarely; Sometimes; Often; Constantly; Not applicable]
\item For the AI tools you use, who pays? [SA: I only use free versions; I pay out of my own pocket; My employer or company provides/pays; My school or university provides/pays; Someone else (e.g., family) pays; I get a student or other discount; Prefer not to say]
\item About how much do you personally pay for AI tools per month, in total? [DD: Less than \$20; \$20--29; \$30--49; \$50--99; \$100--199; \$200--499; \$500 or more; Prefer not to say] \emph{(shown only if ``I pay out of my own pocket'' selected)}
\item About how long have you been paying (out of pocket) for an AI tool? [DD: Less than 3 months; 3--6 months; 6--12 months; 1--2 years; More than 2 years; Prefer not to say] \emph{(shown only if ``I pay out of my own pocket'' selected)}
\end{itemize}
 
\subsection{Task reflections (two blocks, randomized order)}
 
Preface shown before the blocks: \emph{``In this next section, we'll ask about two recent tasks where you used AI --- one where the result DID feel like yours, and one where the result DID NOT feel like yours. These will appear in a random order, so please check each section header before answering.''}
 
Each block began: \emph{``For the next questions, `AI' means tools like ChatGPT, Gemini, Claude, Copilot, or other chatbots and assistants that generate text, answers, images, or recommendations. Please tell us about a task that you recently used AI to complete and where the result [felt very much like yours $\mid$ did not really feel like yours]. Think about whether the result felt like yours, how much you shaped the task direction, and how independently you completed the task. Consider describing a creative project, something you planned for another person, a chore or errand, a decision you made, a work or school project, or anything else that comes to mind. There are no right or wrong answers --- we're just as interested in small, everyday tasks as in big ones.''}
 
\begin{itemize}
\item Briefly, what was the task? [FF]
\item Why did you need to do this task, and how did you use AI to complete it? If you used any other resources to complete the task, please describe them. [FF]
\item In what ways, if any, does the result of this task reflect who you are --- your values, personality, or identity? If it doesn't feel like a reflection of you, why not? [FF]
\item Reflect on other similar tasks you have done. How much does this task feel like your own work? Consider how you did/did not shape the direction of the task, any decisions you made, or how much you relied on others. [FF]
\item Think about how you feel about the outcome of this work. Are you proud of the final result and would you be comfortable telling others you used AI for it? [FF]
\item Was using AI worth it to you? Consider why you chose to use AI for this task and how the result might feel or be different if you hadn't used AI. [FF]
\item Would you still have done this task without AI? [MC: Definitely yes; Probably yes; Might or might not; Probably not; Definitely not]
\item Would you be able to do this task without AI? [MC: Definitely yes; Probably yes; Might or might not; Probably not; Definitely not]
\end{itemize}
 
\subsection{Common tasks with AI}\label{app:commontask}
 
Below are common tasks done using AI. Select any you have used AI to do in the past 30 days (1 month). [SA]
 
\begin{itemize}
\item Looking up or checking specific information; Learning about a topic or how something works; How-to help or step-by-step guidance; Summarizing, explaining, or translating a document, article, or video; Editing or improving something I'd already written; Writing messages, emails, or other communication; Creative writing (stories, poems, fiction); Brainstorming or coming up with ideas; Making or editing images, audio, or video; Cooking, recipes, or meal planning; Health, fitness, beauty, or self-care; Medical or health-related questions; Shopping or product research; Planning something (a trip, an event, a gift, a schedule); Advice on a personal problem, relationship, or how I feel; Coding, math, or data analysis; Casual conversation, games, or roleplay; Something else (please specify) (FF)
\end{itemize}
 
\subsection{Comparison and closing}\label{app:comparison}
 
\begin{itemize}
\item As a reminder, here are the tasks you described using AI for earlier. [Task 1; Task 2 inserted in.] Why did you choose to talk about these tasks and not any of the other tasks you selected above? [FF]
\item Between the two tasks you selected, what made the biggest difference in how much ownership you felt? If they felt similar, tell us that too. [FF]
\item Is there anything else about how you use AI --- or about this survey --- you'd like to share? [FF]
\end{itemize}

\section{Survey Participation}
In Table~\ref{tab:flow}, we report the number of participants who completed each section of our survey. Since all questions were optional, the number of participants $n$ who completed each section varies. The main source of attrition comes from the task reflections section. In our figures and analyses, we report all results available and the number of participants $n$ involved. The median completion time for the survey was 24 minutes.

\begin{table}[H]
\centering\small
\caption{Number of participants who completed each section of the survey.}
\label{tab:flow}
\begin{tabular}{lr}
\toprule
\textbf{Stage} & \textbf{n} \\
\midrule
Opened the survey and consented & 117 \\
Continued past the demographics page & 108 \\
Continued past the AI-usage page (reached the task reflections) & 104 \\
Described at least one task (high-ownership 57; low-ownership 56; both 49) & 64 \\
Completed the common-tasks checklist & 53 \\
Answered ``why these tasks'' / ``biggest difference'' & 50 / 48 \\
Reached the end of the survey & 52 \\
\bottomrule
\end{tabular}
\end{table}

\section{Demographics}
\label{app:demographics}
We collect personal demographics as well as AI usage preferences, patterns, and payments to build additional context to ground our study. In Table~\ref{tab:demographics}, we report the personal demographics of our participants. Due to our sourcing channels of university mailing lists, Slack channels, professional networks, and social media, our sample is biased with a bimodal distribution of young adults and senior professionals, who are predominantly Caucasian and US-centric, well-educated, and experienced with AI in the technology or education sector. This sample is a first step towards understanding trends in ownership in AI-assisted tasks. We encourage future work to apply our framework to a broader and more representative population.

\begin{figure}[H]
    \centering
    \includegraphics[width=0.8\textwidth]{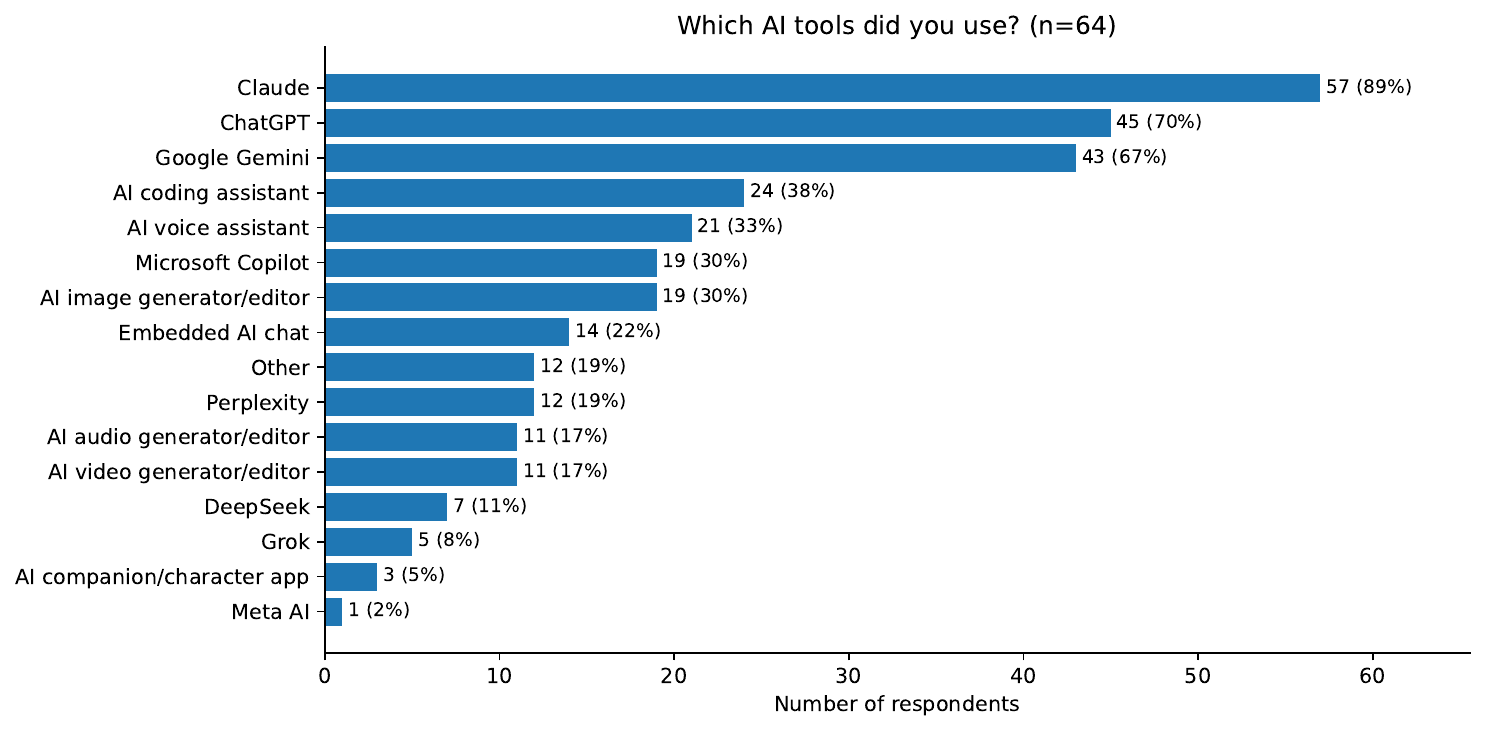}
    \caption{AI tools used by survey respondents (n = 64; select all that apply). Bar labels show the number of respondents who selected each tool, with the percentage of respondents to this question in parentheses. Because participants could select multiple tools, percentages do not sum to 100\%. ``Other'' includes Kimi (x2) and Codex (x2) and 12 other tools mentioned once.}
    \label{fig:ai_tools}
\end{figure}

\begin{figure}[H]
    \centering
    \includegraphics[width=0.8\textwidth]{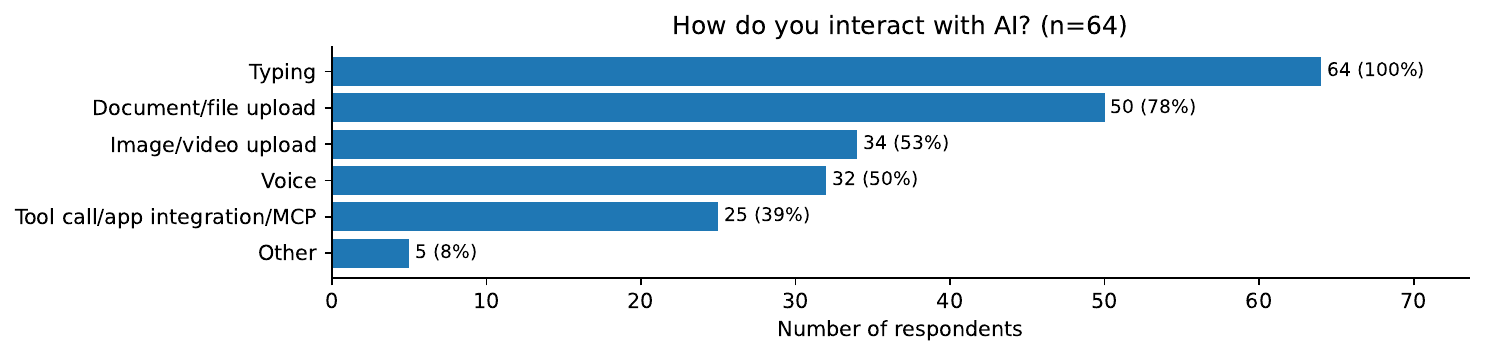}
    \caption{Ways that survey respondents interact with AI tools (n = 64; select all that apply). Bar labels show the number of respondents who selected each tool, with the percentage of respondents to this question in parentheses. Because participants could select multiple tools, percentages do not sum to 100\%. ``Other'' includes API calls (x2), HTML document reviews (x1), and Git (x1).}
    \label{fig:ai_interactions}
\end{figure}

In Table~\ref{tab:ai_usage}, we report AI usage frequency and familiarity from the respondents. Our participants lean towards active users of AI tools and have been using these tools for a while. Figure~\ref{fig:ai_tools} highlights the AI tools participants use. Figure~\ref{fig:ai_interactions} shows the modes of interaction with AI tools. Table~\ref{tab:paid_ai_usage} reveals payment patterns of AI tools.
\subsection{Personal Demographics}

\begin{table}[H]
\centering
\caption{Demographics of survey respondents. All questions were optional, so percentages are calculated based on the number of respondents for each question. For select-all-that-apply items, percentages do not sum to 100\%. Other includes Argentina, New Zealand, Cyprus, Germany, and the Netherlands, each represented by 1 respondent.}
\small
\begin{tabular}{llrr}
\toprule
\textbf{Demographic} & \textbf{Response} & \textbf{n} & \textbf{\%} \\
\midrule
\multirow{6}{*}{\shortstack[l]{\textbf{Age}\\\footnotesize{(n=62)}}}
 & 18–24 & 13 & 21.0\% \\
 & 25–34 & 3 & 4.8\% \\
 & 35–44 & 5 & 8.1\% \\
 & 45–54 & 21 & 33.9\% \\
 & 55–64 & 13 & 21.0\% \\
 & 65 or older & 7 & 11.3\% \\
\midrule
\multirow{6}{*}{\shortstack[l]{\textbf{Country of Residence}\\\footnotesize{(n=60)}}}
 & United States & 43 & 71.7\% \\
 & Australia & 4 & 6.7\% \\
 & Canada & 4 & 6.7\% \\
 & India & 2 & 3.3\% \\
 & United Kingdom & 2 & 3.3\% \\
 & Other* & 5 & 8.3\% \\
\midrule
\multirow{4}{*}{\shortstack[l]{\textbf{Gender}\\\footnotesize{(n=64)}}}
 & Man & 32 & 50.0\% \\
 & Woman & 27 & 42.2\% \\
 & Other & 4 & 6.2\% \\
 & Prefer not to say & 1 & 1.6\% \\
\midrule
\multirow{9}{*}{\shortstack[l]{\textbf{Race/Ethnicity}\\\textit{(select all)}\\\footnotesize{(n=64)}}}
 & American Indian or Alaska Native & 0 & 0.0\% \\
 & Asian & 7 & 10.9\% \\
 & Black or African American & 0 & 0.0\% \\
 & Hispanic or Latino/a/e & 4 & 6.2\% \\
 & Middle Eastern or North African & 0 & 0.0\% \\
 & Native Hawaiian or Other Pacific Islander & 0 & 0.0\% \\
 & White or Caucasian & 51 & 79.7\% \\
 & Other & 1 & 1.6\% \\
 & Prefer not to say & 5 & 7.8\% \\
\midrule
\multirow{8}{*}{\shortstack[l]{\textbf{Education}\\\footnotesize{(n=64)}}}
 & Less than high school & 0 & 0.0\% \\
 & High school / GED & 2 & 3.1\% \\
 & Some college & 14 & 21.9\% \\
 & Associate's degree & 0 & 0.0\% \\
 & Bachelor's degree & 17 & 26.6\% \\
 & Master's degree & 18 & 28.1\% \\
 & Doctorate / Professional degree & 12 & 18.8\% \\
 & Prefer not to say & 1 & 1.6\% \\
\midrule
\multirow{6}{*}{\shortstack[l]{\textbf{Employment Status}\\\footnotesize{(n=63)}}}
 & Employed full-time & 25 & 39.7\% \\
 & Employed part-time & 3 & 4.8\% \\
 & Self-employed & 18 & 28.6\% \\
 & Student & 11 & 17.5\% \\
 & Not currently employed & 1 & 1.6\% \\
 & Retired & 5 & 7.9\% \\
\midrule
\multirow{11}{*}{\shortstack[l]{\textbf{Work Industry}\\\footnotesize{(n=63)}}}
 & Education & 10 & 15.9\% \\
 & Finance / insurance & 1 & 1.6\% \\
 & Government / public sector & 2 & 3.2\% \\
 & Healthcare & 3 & 4.8\% \\
 & Legal & 3 & 4.8\% \\
 & Media / arts / design & 3 & 4.8\% \\
 & Retail / hospitality & 1 & 1.6\% \\
 & Science / engineering & 4 & 6.3\% \\
 & Technology / software & 20 & 31.7\% \\
 & Other & 8 & 12.7\% \\
 & Not applicable & 8 & 12.7\% \\
\bottomrule
\end{tabular}
\label{tab:demographics}
\end{table}

\subsection{AI Usage}

\begin{table}[H]
\centering
\caption{Self-reported AI usage among survey respondents. Percentages are calculated based on the number of respondents for each question, as not all participants answered every question.}
\small
\begin{tabular}{llrr}
\toprule
\textbf{Question} & \textbf{Response} & \textbf{n} & \textbf{\%} \\
\midrule
\multirow{6}{*}{\shortstack[l]{\textbf{Usage Frequency (Past 3 Months)}\\\footnotesize{(n=54)}}}
 & Many times a day & 35 & 64.8\% \\
 & Daily & 5 & 9.3\% \\
 & Several times a week & 8 & 14.8\% \\
 & Weekly & 2 & 3.7\% \\
 & Monthly & 2 & 3.7\% \\
 & Less than monthly & 2 & 3.7\% \\
\midrule
\multirow{6}{*}{\shortstack[l]{\textbf{Length of AI Use}\\\footnotesize{(n=64)}}}
 & More than 2 years ago & 30 & 46.9\% \\
 & 1–2 years ago & 12 & 18.8\% \\
 & 6–12 months ago & 8 & 12.5\% \\
 & 3–6 months ago & 2 & 3.1\% \\
 & In the last 3 months & 8 & 12.5\% \\
 & I don't use them regularly & 4 & 6.2\% \\
\midrule
\multirow{6}{*}{\shortstack[l]{\textbf{AI Use in Job}\\\footnotesize{(n=57)}}}
 & Constantly & 28 & 49.1\% \\
 & Often & 12 & 21.1\% \\
 & Sometimes & 6 & 10.5\% \\
 & Rarely & 4 & 7.0\% \\
 & Not at all & 2 & 3.5\% \\
 & Not applicable & 5 & 8.8\% \\
\bottomrule
\end{tabular}
\label{tab:ai_usage}
\end{table}

\subsection{Paid AI Usage}

\begin{table}[H]
\centering
\caption{Self-reported paid AI usage among survey respondents. Percentages are calculated based on the number of respondents for each question, as not all participants answered every question. For select-all-that-apply items, percentages do not sum to 100\%. The Monthly Spend and Length of Paid Use questions are restricted to respondents who reported paying for AI usage out of pocket.}
\small
\begin{tabular}{llrr}
\toprule
\textbf{Question} & \textbf{Response} & \textbf{n} & \textbf{\%} \\
\midrule
\multirow{5}{*}{\shortstack[l]{\textbf{Who Pays} \textit{(select all)}\\\footnotesize{(n=63)}}}
 & I only use free versions & 17 & 27.0\% \\
 & I pay out of my own pocket & 36 & 57.1\% \\
 & I get a student or other discount & 4 & 6.3\% \\
 & My employer or company provides/pays & 33 & 52.4\% \\
 & My school or university provides/pays & 6 & 9.5\% \\
\midrule
\multirow{8}{*}{\shortstack[l]{\textbf{Monthly Spend (\$)}\\\footnotesize{(n=36)}}}
 & Less than \$20 & 6 & 16.7\% \\
 & \$20-29 & 2 & 5.6\% \\
 & \$30-49 & 8 & 22.2\% \\
 & \$50-99 & 5 & 13.9\% \\
 & \$100-199 & 7 & 19.4\% \\
 & \$200-499 & 6 & 16.7\% \\
 & \$500 or more & 1 & 2.8\% \\
 & Prefer not to say & 1 & 2.8\% \\
\midrule
\multirow{5}{*}{\shortstack[l]{\textbf{Length of Paid Use}\\\footnotesize{(n=36)}}}
 & More than 2 years & 14 & 38.9\% \\
 & 1–2 years & 8 & 22.2\% \\
 & 6–12 months & 10 & 27.8\% \\
 & 3–6 months & 2 & 5.6\% \\
 & Less than 3 months & 2 & 5.6\% \\
\bottomrule
\end{tabular}
\label{tab:paid_ai_usage}
\end{table}

\section{Survey Results}
\subsection{Interest and Ability in Performing Tasks without AI}
In Figure~\ref{fig:front}, we highlight the difference between participants being \textit{able} to do the task without AI across high and low ownership tasks, which reveals a difference among those who responded ``definitely yes''. When compared to whether they would have done the task without AI, there is less of a difference between high and low ownership tasks.

\begin{figure}[H]
    \centering
    \includegraphics[width=\textwidth]{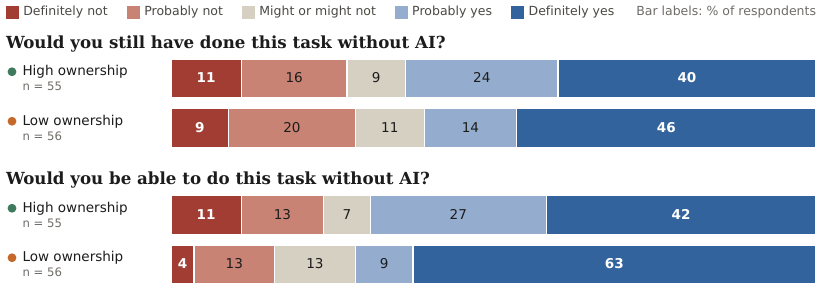}
    \caption{Responses to questions about task completion without AI: whether respondents would have done the task (top) and whether they would have \textit{been able to} do the task (bottom). Within each subplot, responses are split by high ownership and low ownership. Bars show the proportion of respondents in each group selecting each of the five Likert options \textit{(Definitely yes, Probably yes, Might or might not, Probably not, Definitely not)}.}
    \label{fig:ai_ability}
\end{figure}

\subsection{Pride and Disclosure}
When asked about whether they felt proud of the result, respondents are generally proud of high ownership tasks and mixed for low ownership tasks. Disclosure of AI usage remains similar across high and low ownership tasks, though 10\% are uncomfortable disclosing AI use for low ownership tasks. When pooled across high and low ownership tasks, comfort with disclosure is loosely tied to pride.
\label{app:disclosure}
\begin{figure}[H]
    \centering
    \includegraphics[width=0.9\textwidth]{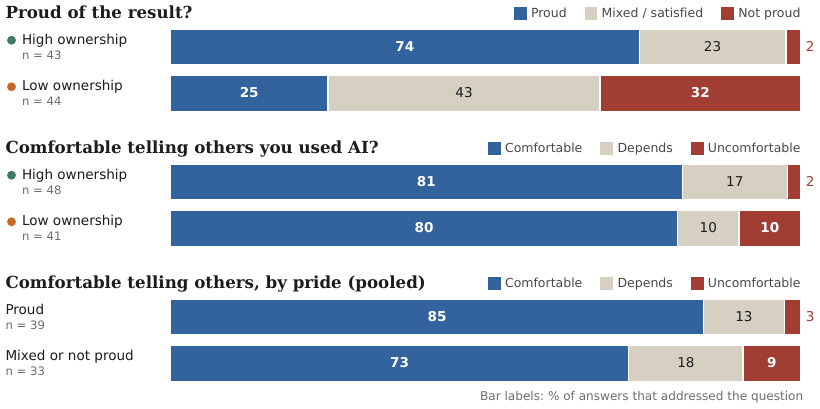}
    \caption{Answers to ``Are you proud of the final result and would you be comfortable telling others you used AI for it?'', coded by fixed keyword rules (proud / mixed or merely satisfied / not proud; comfortable / depends on audience, stakes or being able to explain the extent / uncomfortable). Bottom panel: both tasks pooled, comparing pride and disclosure trends ($n = 72$).}
    \label{fig:pride_disclosure}
\end{figure}

\subsection{Common Tasks and AI Memory Gap}

\begin{figure}[H]
    \centering
    \includegraphics[width=\textwidth]{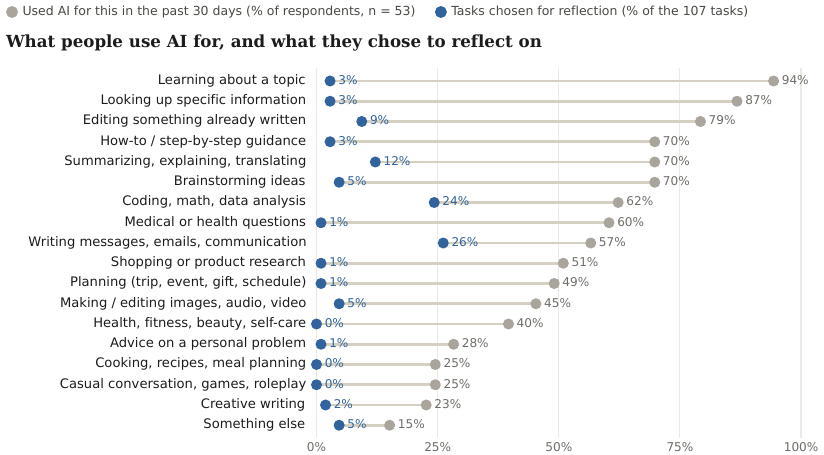}
    \caption{Everyday AI use versus the tasks chosen for reflection. Grey: percent of checklist respondents ($n = 53$) who used AI for each category in the past 30 days. Blue: percent of the 107 reflected tasks (mapped to checklist category via rule-based coding and LLM judging).}
    \label{fig:memory_gap}
\end{figure}

On the common tasks checklist (Appendix~\ref{app:commontask}), respondents ($n = 53$) reported a median of 9 of 17 task categories in the past 30 days. The most common were learning about a topic (94\%) and looking up specific information (87\%). Of the tasks respondents chose to reflect on, only a handful were lookup- or learning-related tasks. Tasks respondents chose to reflect on were often writing and technical tasks. Respondents' explanations for their choices (Appendix~\ref{app:comparison}) were due to recency bias, salience, and having a clear result to judge ownership of. Ownership reflections often involved tasks that produced a significant artifact rather than informational use from day-to-day interaction. This is consistent with the literature on people underestimating their AI usage \citep{yu2026cognitive} and forgetting what they created with AI \citep{aimemorygap}.

\subsection{Anecdotes on Tasks}

\begin{table}[H]
\centering\footnotesize
\caption{Pairs of high and low ownership tasks described by the same respondent, with the respondent's own explanation of the difference. The tag under each task is that respondent's answer to ``Would you be able to do this task without AI?''.}
\label{tab:anecdotes}
\renewcommand{\arraystretch}{1.15}
\begin{tabular}{>{\raggedright\arraybackslash}p{1.9cm}>{\raggedright\arraybackslash}p{3.2cm}>{\raggedright\arraybackslash}p{3.2cm}>{\raggedright\arraybackslash}p{4.0cm}}
\toprule
\textbf{Respondent} & \textbf{High ownership} & \textbf{Low ownership} & \textbf{What's the difference?} \\
\midrule
Tech executive & Graphic novels adapted from their own novels\newline{\textcolor[HTML]{a13d33}{\rule[-0.1ex]{0.75em}{0.75em}}}\,{\scriptsize\textcolor[HTML]{6f6d68}{definitely not}} & Negotiating an Amazon refund through an AI agent\newline{\textcolor[HTML]{33639c}{\rule[-0.1ex]{0.75em}{0.75em}}}\,{\scriptsize\textcolor[HTML]{6f6d68}{definitely yes}} & ``I didn't draw the art, but I feel the same as if I would have hired a human artist'' \\
Solo-practice attorney & An AI pipeline that files scanned documents to the right matter\newline{\textcolor[HTML]{a13d33}{\rule[-0.1ex]{0.75em}{0.75em}}}\,{\scriptsize\textcolor[HTML]{6f6d68}{definitely not}} & A memorandum supporting an order to show cause\newline{\textcolor[HTML]{33639c}{\rule[-0.1ex]{0.75em}{0.75em}}}\,{\scriptsize\textcolor[HTML]{6f6d68}{definitely yes}} & ``I rewrite all communication generated by AI so I own it'' \\
Research mentor & A team of agents building speech-recognition software\newline{\textcolor[HTML]{a13d33}{\rule[-0.1ex]{0.75em}{0.75em}}}\,{\scriptsize\textcolor[HTML]{6f6d68}{definitely not}} & Directing a student paper that students revised with AI\newline{\textcolor[HTML]{33639c}{\rule[-0.1ex]{0.75em}{0.75em}}}\,{\scriptsize\textcolor[HTML]{6f6d68}{definitely yes}} & ``Who's doing the prompting, who's defending the claims'' \\
IT analyst and digital artist & A 3-D multichannel audio-synthesis environment\newline{\textcolor[HTML]{a13d33}{\rule[-0.1ex]{0.75em}{0.75em}}}\,{\scriptsize\textcolor[HTML]{6f6d68}{definitely not}} & An adventure game about personal issues, built hands-off\newline{\textcolor[HTML]{a13d33}{\rule[-0.1ex]{0.75em}{0.75em}}}\,{\scriptsize\textcolor[HTML]{6f6d68}{definitely not}} & ``Task 2 was intended to be as hands off as possible, so I would not call the results mine'' \\
Artisan & Diagnosing a car problem while steering the search\newline{\textcolor[HTML]{c98374}{\rule[-0.1ex]{0.75em}{0.75em}}}\,{\scriptsize\textcolor[HTML]{6f6d68}{probably not}} & Diagnosing a car problem while passively receiving the answer\newline{\textcolor[HTML]{c98374}{\rule[-0.1ex]{0.75em}{0.75em}}}\,{\scriptsize\textcolor[HTML]{6f6d68}{probably not}} & ``how I engaged in the direction of the search'' \\
Retiree & Planning an overseas trip through extended back-and-forth\newline{\textcolor[HTML]{94adce}{\rule[-0.1ex]{0.75em}{0.75em}}}\,{\scriptsize\textcolor[HTML]{6f6d68}{probably yes}} & Identifying a bird from a photo\newline{\textcolor[HTML]{d6d0c2}{\rule[-0.1ex]{0.75em}{0.75em}}}\,{\scriptsize\textcolor[HTML]{6f6d68}{might or might not}} & ``a long process with much back and forth. I think that made me feel more ownership'' \\
Student (research intern) & Debugging cluster software by talking through the errors\newline{\textcolor[HTML]{94adce}{\rule[-0.1ex]{0.75em}{0.75em}}}\,{\scriptsize\textcolor[HTML]{6f6d68}{probably yes}} & A school-club website\newline{\textcolor[HTML]{c98374}{\rule[-0.1ex]{0.75em}{0.75em}}}\,{\scriptsize\textcolor[HTML]{6f6d68}{probably not}} & ``Claude was as confused as me, so when the problem finally did get solved I felt like I actually played a role'' \\
Student & Personal blog posts\newline{\textcolor[HTML]{94adce}{\rule[-0.1ex]{0.75em}{0.75em}}}\,{\scriptsize\textcolor[HTML]{6f6d68}{probably yes}} & A GIS data-processing pipeline\newline{\textcolor[HTML]{c98374}{\rule[-0.1ex]{0.75em}{0.75em}}}\,{\scriptsize\textcolor[HTML]{6f6d68}{probably not}} & ``The \% of text created that was typed by me rather than copy-and-pasted'' \\
Postdoctoral researcher & Streaming camera data from an embedded device for robot training\newline{\textcolor[HTML]{94adce}{\rule[-0.1ex]{0.75em}{0.75em}}}\,{\scriptsize\textcolor[HTML]{6f6d68}{probably yes}} & A conference website\newline{\textcolor[HTML]{33639c}{\rule[-0.1ex]{0.75em}{0.75em}}}\,{\scriptsize\textcolor[HTML]{6f6d68}{definitely yes}} & ``The first task I did on my own volition and self-interest, the second task I did because it was a responsibility'' \\
Engineer & A map of locations for a website\newline{\textcolor[HTML]{94adce}{\rule[-0.1ex]{0.75em}{0.75em}}}\,{\scriptsize\textcolor[HTML]{6f6d68}{probably yes}} & A comedic letter about a friend, read at a party\newline{\textcolor[HTML]{33639c}{\rule[-0.1ex]{0.75em}{0.75em}}}\,{\scriptsize\textcolor[HTML]{6f6d68}{definitely yes}} & ``The strength of my initial vision for what the result should be, and the amount of existing work I had before I used the tool'' \\
Clinical informatics specialist & Deciding which jobs to target in a career change\newline{\textcolor[HTML]{d6d0c2}{\rule[-0.1ex]{0.75em}{0.75em}}}\,{\scriptsize\textcolor[HTML]{6f6d68}{might or might not}} & A LinkedIn community post\newline{\textcolor[HTML]{33639c}{\rule[-0.1ex]{0.75em}{0.75em}}}\,{\scriptsize\textcolor[HTML]{6f6d68}{definitely yes}} & ``the AI helped with the pattern recognition that I wasn't seeing in my own experience and choices'' \\
QA analyst & Test-coverage tracking and documentation\newline{\textcolor[HTML]{33639c}{\rule[-0.1ex]{0.75em}{0.75em}}}\,{\scriptsize\textcolor[HTML]{6f6d68}{definitely yes}} & Formatting test cases\newline{\textcolor[HTML]{33639c}{\rule[-0.1ex]{0.75em}{0.75em}}}\,{\scriptsize\textcolor[HTML]{6f6d68}{definitely yes}} & ``the ownership felt the same. It was work that I did with the use of the tool'' \\
\bottomrule
\end{tabular}
\end{table}



\end{document}